\documentclass{article}
\pdfoutput=1

\usepackage{arxiv}

\usepackage[utf8]{inputenc} 
\usepackage[T1]{fontenc}    
\usepackage{hyperref}       
\usepackage{url}            
\usepackage{booktabs}       
\usepackage{amsfonts}       
\usepackage{nicefrac}       
\usepackage{microtype}      
\usepackage{graphicx}
\usepackage{lipsum}
\usepackage{amsmath}
\usepackage{tabularx}
\usepackage{ctable} 
\usepackage{wrapfig}
\usepackage[abs]{overpic}
\usepackage{pict2e}
\usepackage[backend=biber, style=ieee, maxnames=5, minnames=3]{biblatex}
\DeclareMathOperator*{\argmin}{arg\,min}
\DeclareMathOperator{\arctan2}{arctan2}

\DeclareMathOperator*{\EE}{\mathbb E\,}
\newcommand{\ATA}{\textup{\textbf{AT+A}}}
\newcommand{\ATTIA}{\textup{\textbf{AT+TI+A}}}
\newcommand{\RTTIA}{\textup{\textbf{RT+TI+A}}}
\newcommand{\ATTI}{\textup{\textbf{AT+TI}}}
\newcommand{\ATTIU}{\textup{\textbf{AT+TI/U}}}
\newcommand{\ATrcs}{\textup{\textbf{AT,\;rcs}}}

\newcommand{\beginsupplement}{%
        \setcounter{table}{0}
        \renewcommand{\figurename}{Supplementary Figure}
        \setcounter{figure}{0}
        \renewcommand{\tablename}{Supplementary Table}
     }

\title{Novel tracking approach based on fully-unsupervised disentanglement of the geometrical factors of variation}

\author{
  Mykhailo Vladymyrov\thanks{Presently at: Theodor Kocher Institute, University of Bern and Science IT Support, University of Bern} \\
  Albert Einstein Center for Fundamental Physics\\
  Laboratory for High Energy Physics\\
  University of Bern\\
  Bern 3012, Switzerland \\
  \texttt{mykhailo.vladymyrov@tki.unibe.ch} \\
   \And
 Akitaka Ariga \\
  Albert Einstein Center for Fundamental Physics\\
  Laboratory for High Energy Physics\\
  University of Bern\\
  Bern 3012, Switzerland \\
  \texttt{akitaka.ariga@lhep.unibe.ch} \\
}

\begin{document}
\maketitle

\begin{abstract}
Efficient tracking algorithms are a crucial part of particle tracking detectors. While a lot of work has been done in designing a plethora of algorithms, these usually require tedious tuning for each use case. (Weakly) supervised Machine Learning-based approaches can leverage the actual raw data for maximal performance. Yet in realistic scenarios, sufficient high-quality labeled data is not available. While training might be performed on simulated data, the reproduction of realistic signal and noise in the detector requires substantial effort, compromising this approach. \\
Here we propose a novel, fully unsupervised, approach to track reconstruction. The introduced model for learning to disentangle the factors of variation in a geometrically meaningful way employs geometrical space invariances. We train it through constraints on the equivariance between the image space and the latent representation in a Deep Convolutional Autoencoder. Using experimental results on synthetic data we show that a combination of different space transformations is required for meaningful disentanglement of factors of variation. We also demonstrate the performance of our model on real data from tracking detectors.
\end{abstract}


\section{Introduction}
\label{sec:inro}
Particle tracking detectors allow us to study elementary particle interactions by visualizing particle trajectories.
Robust tracking algorithms are nowadays a fundamental component of all tracking detector techniques. Tracking techniques in particle physics have evolved along with technological developments, from implementations on hardware logical elements, computer data processing, GPU-accelerated algorithms, to modern Deep-Learning based approaches \cite{NiwaK1974,Alexandrov2017,Acciarri2017}. The advanced implementation of tracking algorithms can be seen for example in emulsion detector data reconstruction.

\begin{figure}
    \centering
    \includegraphics[width=5cm]{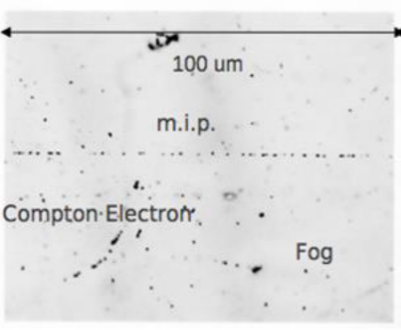}
    
    \caption{An emulsion detector viewed under a microscope. Tracks of charged particles are visible as sequences of silver grains. The detector is sensitive to particles with minimal ionization (m.i.p.).}
    \label{fig:fig1a}
\end{figure}

Nuclear photoemulsion (referred to as \textit{emulsion} in further text) detectors are tracking detectors that allow the detection of charged particles with high spatial (50 nm) and angular (<1 mrad) resolution. They do not require a power supply during the experimental run. These properties enable fundamental physical experiments searching for short lived particles \cite{Kodama2001, Agafonova2015, Ahdida2019, Aghion2018} and large scale experiments in remote regions \cite{Nishiyama2017}. The emulsion gel consists of small silver bromide crystals dispersed in a gelatin frame. When a charged particle passes through the emulsion gel, the crystals along its trajectory create latent image centers, which become visible under optical microscopes after chemical development (Figure \ref{fig:fig1a}).

Conventional track reconstruction is performed in three steps. First, 3D tomographic images of the emulsion detector are acquired using automated scanning microscopes. Next, the positions of silver grains (“hits”) are located in the 3D image volumes, and finally tracks in the detector volume are reconstructed as a sequence of hits along straight lines \cite{Alexandrov2017, Ariga2018}. For this detector, the typical track curvature radius is significantly larger than the track length in a single emulsion film, and thus the local curvature is ignored.

Several tracking algorithms were developed during the evolution of the scanning systems, allowing for efficient track reconstruction in real-time during data acquisition \cite{Armenise2005, Ariga2014, Yoshimoto2017}, as well as for particle identification and energy measurements \cite{Radovic2018,Adams2019,Domine2019}. While satisfying the needs of many experiments they have several drawbacks. Their adaptation to different experimental condition, e.g. high track density or high background level requires tedious calibration ranging from extensive parameter tuning to performing dedicated test runs using e.g. accelerator beams. In addition, when the procedure of extracting the hits is separated from track reconstruction, the tracking algorithm cannot fully exploit the information available in the raw image data, compromising performance especially in the high background/track density cases.

Incorporating tracking based on classical Deep Learning, where the track parameters are predicted from the raw image data, would naturally address the latter issue. Yet, to train such a model in a supervised manner either one would need to provide massive amounts of labeled 3D raw image data for each experimental case, or training would need to be performed largely on simulated datasets. While suitable for some similar cases \cite{Acciarri2017} this approach requires perfect knowledge of the optical microscope parameters, grain size distributions, detector noise, etc., which are not always available directly. 

Similarly to recent works where, for example, the underlying factors of variation in the images of faces, such as eyes or hair color, glasses, or head tilt are disentangled \cite{Kumar2017}, it is possible to identify geometrical factors of variation of tracks. Training such models in an unsupervised manner, i.e. where no track parameters (labels) are provided during the training can address the issues mentioned above simultaneously, by both leveraging raw image data for efficient track reconstruction and allowing simple adaptation to new configurations requiring the raw image dataset only.

In this work we aim at studying an unsupervised learning approach for extracting track parameters solely from the raw image data. Here we introduce a tracking approach based on the Deep Convolutional Autoencoder \cite{Lecun1998, Li2018} model that learns to disentangle the geometrical factors of variation (coordinates and angles of each track) in a fully unsupervised manner by imposing equivariance of the space transformation. While the reconstruction constraint alone fails to disentangle the factors of variation in a meaningful way, we show that adding a simple constraint on the translational invariance along the track line also does not lead to the desired disentanglement. We demonstrate that incorporating more sophisticated transformations in the latent representation is necessary to avoid the reference ambiguity.

The remaining of the paper is structured as follows. In Section \ref{sec:Methods}. the details of the proposed equivariance constraints, latent representation interpretation and implementation details will be given. In Section \ref{sec:Results}. we will show how different constraints affect the performance and carry out an in-depth study of encoder and decoder performance separately to better understand the learned representation. Also, performance on real emulsion data will be shown. Finally, in Section \ref{sec:Discussion} the applicability of the approach and future prospects will be discussed.

\section{Methods}
\label{sec:Methods}
\subsection{Equivariance constraint}
In this work, we will simplify the problem to the 2D case. We use synthetic image data of the emulsion detector tracks to perform a study of the proposed approach. Also, we demonstrate the performance of the trained model on real emulsion detector data.
\begin{figure}
    \centering
    \includegraphics[width=10cm]{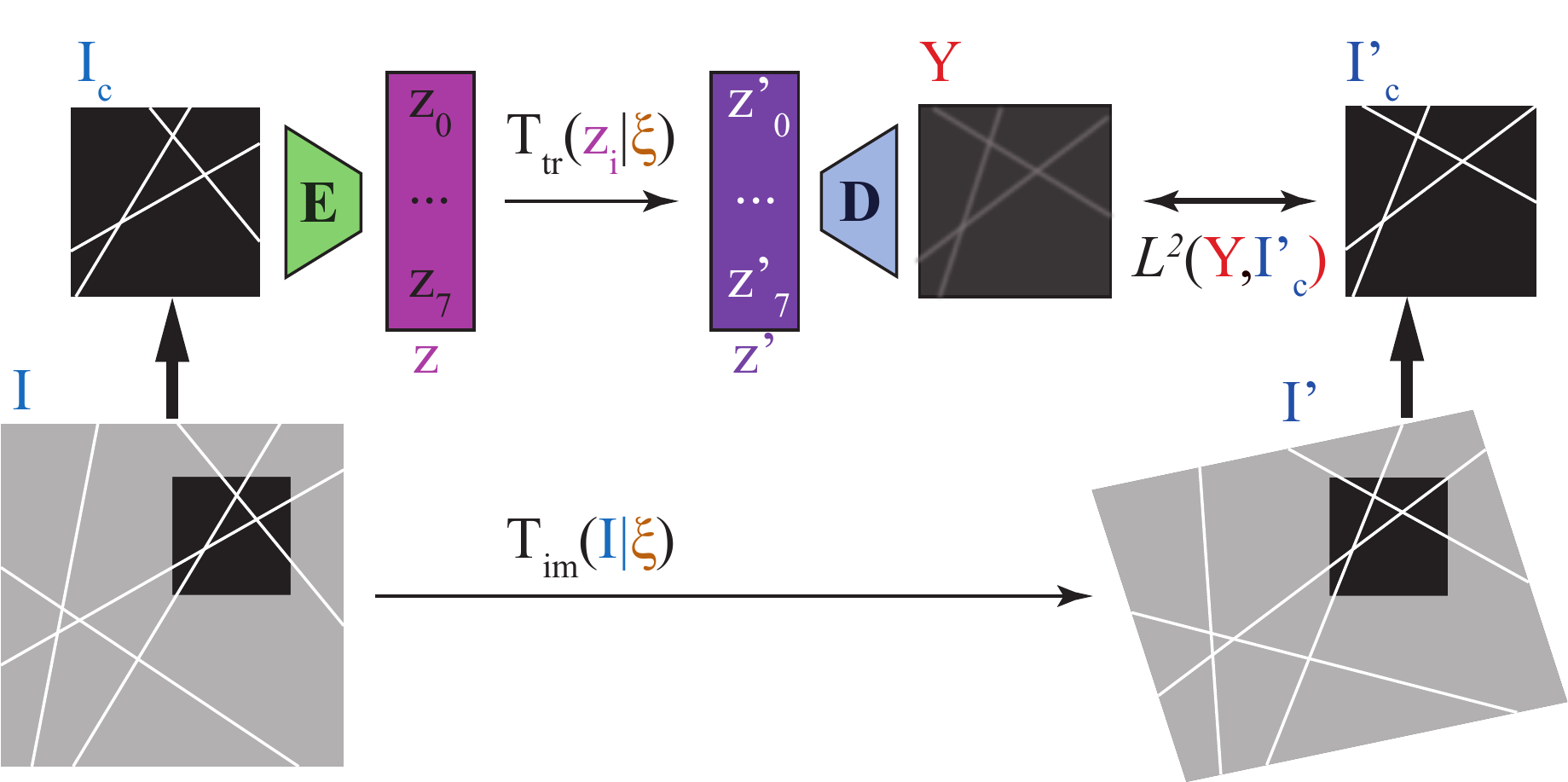}
    
    \caption{Schematic view of Deep Convolutional Autoencoder consisting of an encoder $E$ and a decoder $D$. Geometrical space transformations are applied to images $I$ and each of the latent representation blocks $z_i$.}
    \label{fig:fig1b}
\end{figure}

We use a Deep Convolutional Autoencoder consisting of an encoder $E$ and a decoder $D$ as illustrated in Fig. \ref{fig:fig1b}, that are trained in an end-to-end manner. The encoder $E$ acts on 32$\times$32 pixel images $I_c$, which are obtained from the full images $I$ using the cropping operation $C$: $I_c=C(I)$, producing the latent representation, $z$. In our setup, $z$ is used to estimate the geometrical parameters, namely position and angle, of tracks present in the image crop.

We then define a set of geometrical transformations acting both in the image representation space and in the latent representation space parametrized by the same parameter set $\xi$. In the image space $I'= T_{im}(I \vert \xi)$ and in the latent space of track parameters $z'_t=T_{tr} (z_t \vert \xi)$.

Given the encoder and decoder functions
$$z=E(I_c ),$$
$$Y=D(z),$$
we then demand equivariance of both encoder and decoder under these transformations, i.e. the commutation of the encoder and decoder functions $E$ and $D$ with the transformations $T$ in corresponding domain:
$$T_{tr} (E(I) \vert \xi)=(E(T_{im}(I \vert \xi)); \forall \xi$$
$$D(T_{tr} (z \vert \xi))=T_{im} (D(z) \vert \xi); \forall \xi$$

From which, assuming $I=D(E(I))$, it follows that
$$D(T_{tr} (E(I) \vert \xi))=T_{im} (I \vert \xi); \forall\xi,$$
where cropping operations are omitted for brevity. This allows us to formulate the optimization problem in an end-to-end manner, primarily through the minimization of the $L^2$ loss between the cropped transformed image $I_c'=C(I')$ and the decoder output $Y$ (Figure \ref{fig:fig1b}):
$$E,D=\argmin_{E,D} \EE_{I,\xi} L^2(Y,I'_c)=
\argmin_{E,D} \EE_{I,\xi} L^2(D(T_{tr} (E(C(I)) \vert \xi)), C(T_{im} (I \vert \xi))).$$
We will show that with a sufficient set of transformations $T$, the model is able to learn a geometrically meaningful latent representation $E(I)$.

\subsection{Interpretable latent representation}
We limit the number of tracks potentially detected on each cropped image to $n=8$, slightly above the maximum possible number of tracks per crop (five) in our data set (see section \ref{sec:trainingdata} for details). We will parametrize a track with $n_p$ parameters. Thus, the encoder is designed to output a vector $z$ of length $n \cdot n_p$. This vector is then partitioned into $n$ chunks of length $n_p$. We further refer to these chunks as “track feature containers” $z_i$, each corresponding to one of the $n$ tracks.
\begin{figure}
    \centering
    \includegraphics[width=5cm]{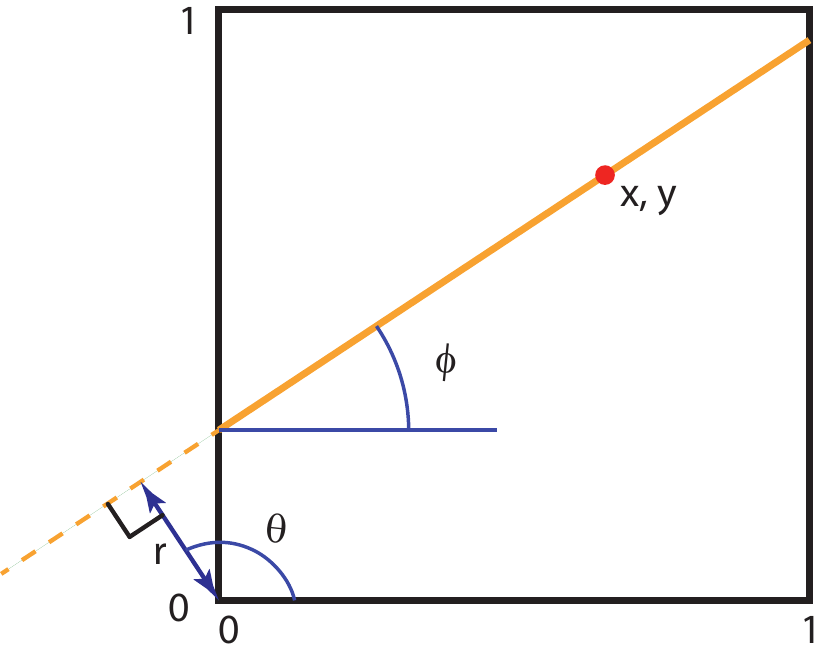} 
    
    \caption{Coordinate range on the image is chosen to be $(0,0)$ at bottom left to $(1,1)$ at right top. Track line is parametrized either by a point on the track $(x,y)$ and sine and cosine of its slope angle $\phi$ as $c=\kappa \cos(\phi)$, $c=\kappa \sin(\phi)$, or by distance $r$ to the track from the origin and angle $\theta$ as $c=\kappa \cos(\theta)$, $c=\kappa \sin(\theta)$.}
    \label{fig:fig1c}
\end{figure}

We attribute an \textit{a priori} meaning to each of the $n_p$ elements in track features $z_i$. Here we have explored three parametrization options:
\begin{enumerate}
    \item $n_p=4$, $z_i=(x_i,y_i,c_i,s_i)$ – the track’s geometrical parameters, where $x_i$, $y_i$ are the coordinates of a point on the $i$-th track line and $c_i=\kappa \cos \phi_i$, $s_i=\kappa \sin \phi_i$ are proportional to the cosine and sine of the track inclination angle $\phi_i$. Coordinates on the image crop are $(0,0)$ in the lower left corner and $(1,1)$ in the upper right, $\phi\in[0, 2\pi)$, so that $\sin \phi, \cos \phi \in[-1,1]$ (Figure \ref{fig:fig1c}). Such a parametrization is chosen because it is continuous and confined, unlike e.g. $\phi$ itself or $\tan \phi$.
    
    \item $n_p=5$, $z_i=(x_i,y_i,c_i,s_i,a_i)$, where the first 4 elements are the track’s geometrical parameters as in 1), and the parameter $a_i \in [0,1]$ shows the confidence of the encoder in the track presence. A value of $a_i \leq 0$ is defined as a disabled track, and $a_i > 0$ as enabled. By disabling some tracks, degenerate outputs, in which multiple containers predict the same track, can be prevented.
	
	\item $n_p=3$, $z_i=(r_i,c_i,s_i)$ – the track’s geometrical parameters in the rho-theta parametrization \cite{Duda1972}. $c_i=\kappa \cos \theta_i$, $s_i=\kappa \sin \theta_i$ – are proportional to the cosine and sine of the angle $\theta_i = \phi_i + \frac {\pi}{2}$, and $r_i$ is the distance from the origin $(0,0)$ to the track (Fig. \ref{fig:fig1c}).
\end{enumerate}

The first two parametrizations are overparametrized yet more naturally occurring in the image representation. Importantly, these enable an explicit implementation of the translation invariance.
The last one is the most common parametrization for 2D tracking (e.g. Hough transformation \cite{Duda1972}). 

Since by implementation the values are $z_i \in [-1,1]$, the parameter $r_i$ is scaled linearly into the  $[0,\sqrt 2]$ range.

\subsection{Representation transformations}
Image transformations are often used for image augmentation during model training \cite{Ronneberger2015} and in some cases, models are trained to recover original images from transformed ones \cite{Jaderberg}. Instead, here we explicitly apply transformations coherently in the image and latent spaces, as required for the equivariance constraint.
As space transformations we employ affine transformations as a combination of rotation, scaling, skew, and translation. Under these affine transformations, straight lines are transformed to straight lines. We implemented these transformations coherently in the image and latent representation spaces. Details on the transformation’s implementation are given in the Appendix \ref{app:Transformations}.

The main property of a line is the translational invariance along it. In the present work we tried to see if this translation transformation can be sufficient for disentangling the geometrical parameters of a line in the latent representation. Also, we have studied the effect of such a transformation when incorporated in addition to the affine transformations.

In this work we included the five model configurations of constraints based on the equivariance between the image space and the latent representation of the track line parameters (Table \ref{table:tab1}).

\makeatletter
\renewcommand\tabularxcolumn[1]{>{\let\Centering}m{#1}}
\makeatother

{\renewcommand{\arraystretch}{1.5}%
    \begin{table}[h!]
     \caption{Model configurations.}
     \label{table:tab1}

     \centering
     \begin{tabularx}{\textwidth}{| c | l | X | c | c |} 
    
     \hline
     & Name & Description & $n_p$ & Parametrization \\
     \specialrule{.1em}{.05em}{.05em}
     1 & \ATA{} & Affine Transformations with track activation parameter $a$ & 5 & $(x_i,y_i,c_i,s_i,a_i)$
     \\ \hline
     2 & \ATTIA{} & Affine Transformations + Translation Invariance with track activation parameter $a$ & 5 & $(x_i,y_i,c_i,s_i,a_i)$
     \\ \hline
     3 & \RTTIA{} & Translation Invariance + Rotation transformation only with track activation parameter $a$ & 5 & $(x_i,y_i,c_i,s_i,a_i)$
     \\ \hline
     4 & \ATTI{} & Affine Transformations + Translation Invariance & 4 & $(x_i,y_i,c_i,s_i)$
     \\ \hline
     5 & \ATrcs{} & Affine Transformations in the (r,c,s) parametrization & 3 & $(r_i,c_i,s_i)$
     \\
     \hline
     \end{tabularx}
     
    \end{table}
}
In the models, which output the track activation parameter $a$, when track container $z_i$ is marked by the encoder as not active ($a_i \leq 0$), we reset the geometrical parameters of this track to random values. This operation forces the decoder to learn to ignore the disabled track. 

\subsection{Loss function}
The model training is performed by minimizing the loss function. In the models without the track activation parameter $a$, the loss function contains only the image term $L_{im}$, which describes the dissimilarity between the decoder output $Y$ and the transformed image $I'$ with the $L^2$ measure:
$$L= L_{im}= \EE_{pixels} \left(\lambda \ \alpha_{sig} (c_0+I'_c)^2 + (1-\lambda) \  \alpha_{L2} \right)\ (Y-I'_c)^2$$
Here $\EE$ denotes averaging over all image pixels. The $(\lambda \ \alpha_{sig} (c_0+I'_c)^2 + (1-\lambda) \  \alpha_{L2} )$ term scales the loss in the image regions with high signal intensity $I'_c$ in the beginning of training ($\lambda=1$): $\EE \alpha_{sig} (c_0+I'_c)^2 (Y-I'_c)^2$. The coefficient $c_0$ prevents the loss from dropping to zero in low intensity image regions. As the training progresses, $\lambda$ exponentially decreases, and the loss is relaxed to pure $L^2$: $\EE \alpha_{L2} (Y-I'_c)^2$.

In the models with the track activation parameter $a$, the loss function contains three terms:
$$L= L_{im}+L_{unif\_act}+L_{bin\_act}$$
The first term describes the $L^2$ pixel value measure as described above. The next two terms address the information flow problem (also referred to as shortcut problem in ref. \cite{Szabo2018}), in which the geometrical parameters $(x,y,c,s)_i$ of multiple tracks describe the same track or some of them are ignored. This is achieved in two steps. First, we demand that each track $t_i$ is found (and marked as active by the parameter $a_i>0$) on average at the same rate as the others:
$$L_{unif\_act}  = \alpha_{unif} \EE_i (\bar a_i - \bar a)^2,$$
where the mean activation of all tracks $\bar a=\EE_{mb,i} a_i$ and the mean activation of a particular track $\bar a_i = \EE_{mb} a_i$ are calculated over the minibatch. The final term $L_{bin\_act}=\alpha_{bin} \EE(1 - a^2)$ forces the activation parameter $a$ to cluster at values $-1$ or $1$, enforcing the encoder decision on whether a track is enabled or disabled.

\subsection{Training data}
\label{sec:trainingdata}
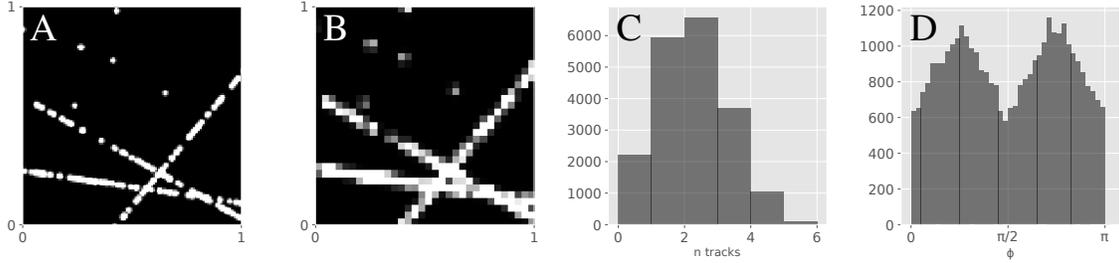
\begin{figure}
\centering
    \begin{overpic}[scale=.365,unit=1mm]{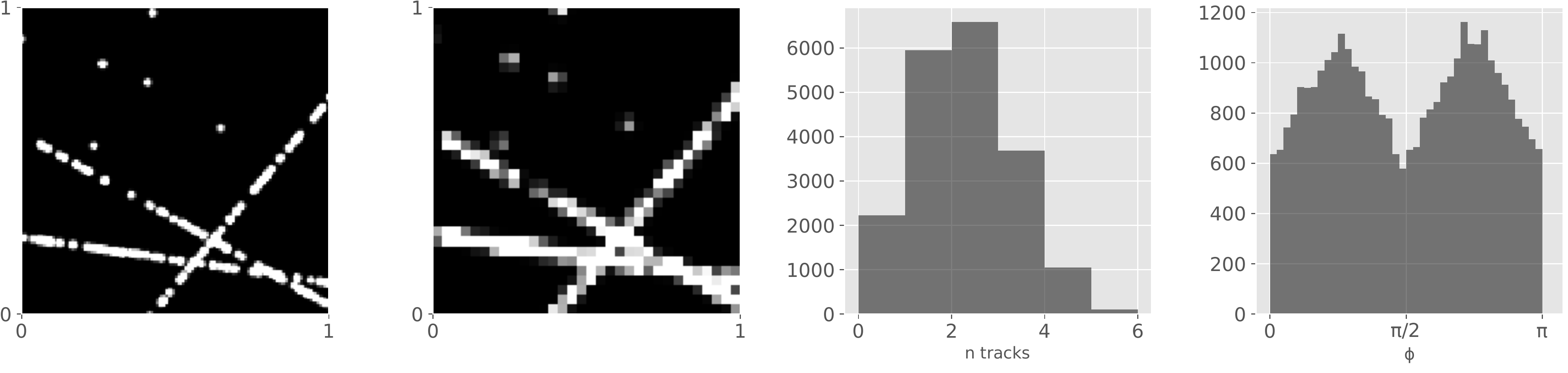} 
        \Large
        \put(3,     30){\color{white} A}  
        \put(42,    30){\color{white} B} 
        \put(81,    30){\color{black} C}
        \put(120,   30){\color{black} D}
    \end{overpic}

    \caption{A. Generated dataset sample. B. Downscaled sample used for model training. C, D. Distributions of number of tracks per image and track angles.}
    \label{fig:fig2}
\end{figure}

For model training and evaluation, we generate synthetic images resembling noisy emulsion data. They contain of two types of objects: “particle tracks” and noise, so called “fog”. Tracks are chains of bright Gaussian spots with the spot density per unit length sampled from a Poisson distribution with mean $\mu$ located randomly along the straight lines with deviation $d \in N(0,\sigma_d)$. Fog is represented by Gaussian spots uniformly distributed in the area with density $\rho$. The track density as well as the $\mu$, $\sigma_d$, and $\rho$ parameters approximately correspond to usual experimental conditions \cite{Ariga2018} and remain fixed throughout this study. While the generated dataset resolution matches the usual imaging resolution, we downscale the images by a factor of four to facilitate this study (Figure \ref{fig:fig2}A,B).

\subsection{Model implementation and training procedure}

In both the encoder and decoder, we incorporated the CoordConv approach \cite{Liu2018} in the first layer, which conceptually fits in our study. In this approach, two additional channels, containing $x$ and $y$ coordinates of pixels in the range of $[0,1]$ correspondingly, are concatenated with input data channels before the first convolutional layer. In practice, we observe that CoordConv improves the performance.

Since we deal with several objects of the same nature, we found it reasonable to apply the decoder $D$ to the track feature containers $z_i$ of each $i$-th track separately and then merge the outputs. To this end we first process each of the containers $z_i$ with the same decoder network that outputs single channel map $Y_i=D(z'_i)$ corresponding to the track $t_i$. Then these images are merged into the final output $Y$ such that for each pixel at coordinates $[k,l]$ in the image $Y_{[k,l]}=\sigma(\max_i Y_{i\,[k,l]})$. Here the sigmoid activation function $\sigma(x)=(1+e^{-x} )^{-1}$ is ensuring that the output pixel values are in the range $[0,1]$. This not only reduces the number of parameters in the decoder, but also simplifies the study of encoder performance, as each $z_i$ has the same structure. Alternatively, shuffling the containers $z_i$ within each sample could be employed. Details on the encoder and decoder architectures are given in Table \ref{table:tab2}.

{\renewcommand{\arraystretch}{1.5}%
    \begin{table}[h!]
    \caption{Description of the encoder and decoder used in our models. conv(kernel size, dilation, \# of channels) - convolution; c\_conv - CoordConv, concatenation with 2 channels of $x$ and $y$ coordinates and convolution; AP - average pooling; FC - fully connected layer; c\_tconv - transposed CoordConv: tiling input up to the target size, concatenation with 2 coordinate channels, and convolution. All convolutions and FC layers are followed by batch normalization \cite{Ioffe2015} and ReLU \cite{Nair:2010:RLU:3104322.3104425} activation, unless otherwise stated.}
    \label{table:tab2}

    \centering
     \begin{tabularx}{0.98\textwidth}{| l | X |} 
    
     \hline
     Encoder & c\_conv(3x3, 1, 16), conv(3x3, 2, 16), conv(3x3, 2, 64), conv(3x3, 2, 128), AP(2x2), conv(1x1, 1, 128), conv(3x3, 2, 512), AP(2x2), conv(1x1, 1, 256), FC(1024), FC(512), FC(128), FC($n \cdot n_p$, $\tanh$)
     \\ \hline
     Decoder & 
     \textit{Single block:} \newline 
     FC(16), c\_tconv(1x1, 32), conv(1x1, 1, 32), conv(1x1, 1, 32), conv(1x1, 1, 16), conv(3x3, 1, 16), conv(3x3, 1, no activation)
     \newline
     
     \textit{Blocks merging:} \newline
     Max(\textit{blocks}), sigmoid activation
     \\
     \hline 
     \end{tabularx}
     
    \end{table}
}

The coefficients in the loss function were chosen empirically to balance the values of its terms: $c_0=0.3$, $\alpha_{sig}=10,000$, $\alpha_{L2}=300$, $\alpha_{unif}=2.5\times10^5$, and $\alpha_{bin}=55$. Training starts with $\lambda=1$, and after $l=100k$ iterations $\lambda$ is exponentially decreased every 5k iterations by a factor of $0.9$, so that $L_{im}$ is relaxed to pure $L^2$ after about 200k iterations:

$$\lambda= 
\begin{cases}
    1, & l<100k\\
    0.9^{\frac{l-100k}{5k}}              & l>100k
\end{cases}$$

\begin{figure}
\centering
\includegraphics[width=0.43\linewidth]{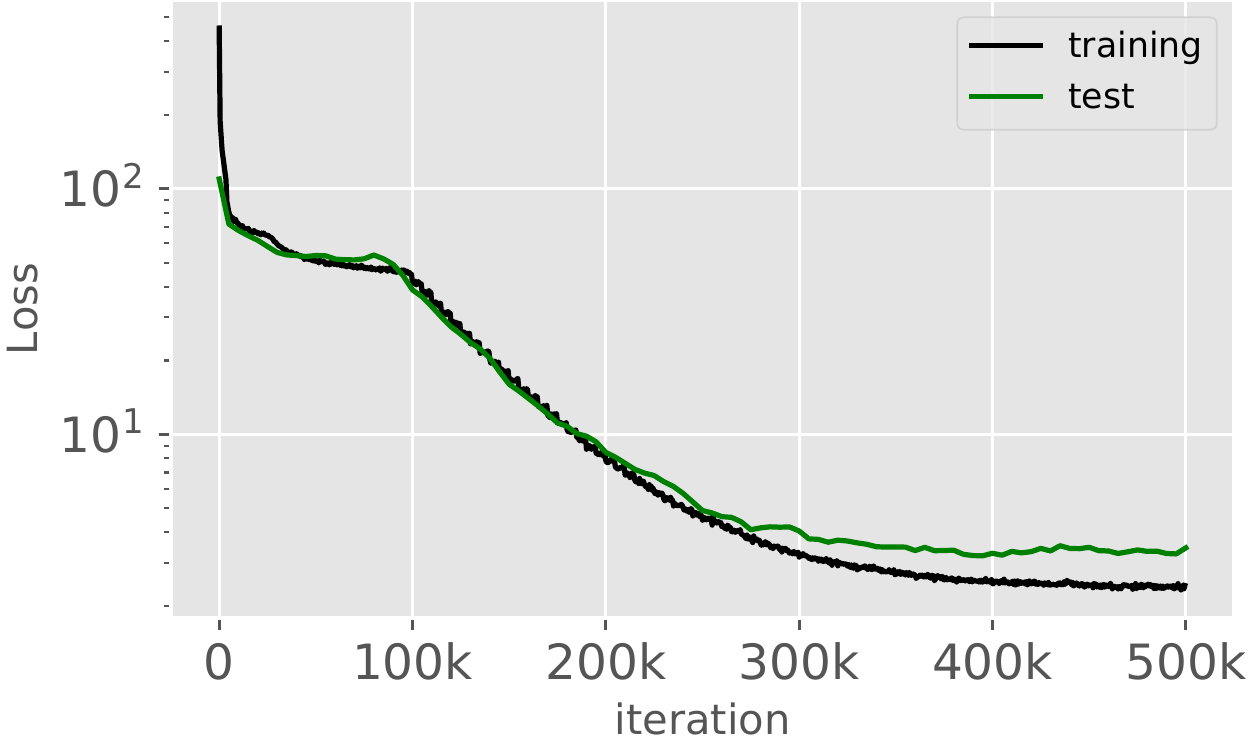} 
\caption{Training and test loss over the course of training of the \ATTI{} model. At iteration 100k the $\lambda$ parameter starts to decrease.}
\label{fig:fig3}
\end{figure}

We perform the training on 32x32 random crops from 40,000 images of 128x128 pixels until convergence for 500,000 iterations with minibatch of 128 images. All experiments were carried out using TensorFlow 1.12 \cite{tensorflow2015-whitepaper}. Models were trained using the Adam optimizer \cite{Kingma2014} with initial learning rate of $6\times10^{-5}$ to allow for stabilization, rising to $1\times10^{-3}$ after 2k iteration was used. Afterwards the rate is decreased by a factor of 0.88 every 90,000 iterations. The loss function over the course of training for e.g. the \ATTI{} model is shown in Figure \ref{fig:fig3}. The loss function evaluated on test dataset (green curve in Fig. \ref{fig:fig3}; see section \ref{sec:traperf} for test dataset details) at training checkpoints confirms that model did not overfit to the training data. The training of each model took about 50 hours on a single GeForce GTX 1080 GPU.

\section{Results}\label{sec:Results}
\subsection{Autoencoder performance}
\begin{figure}
\centering
    \begin{overpic}[trim=-50 0 45 0,clip,width=\textwidth]{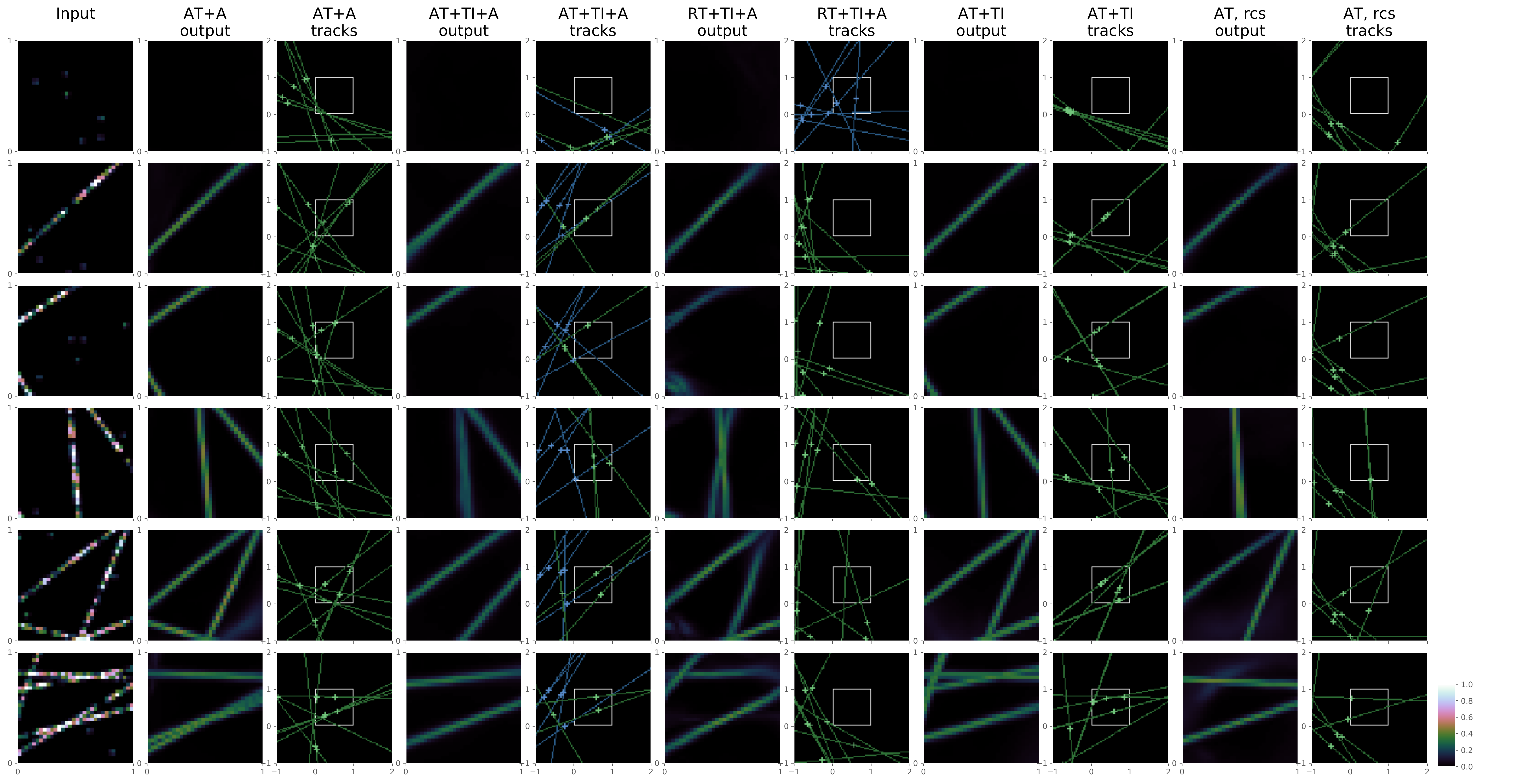} 
        \Large
        \put(2, 210){\color{black} A}  
        \put(2, 171.6){\color{black} B} 
        \put(2, 133.2){\color{black} C}
        \put(2, 94.8 ){\color{black} D}
        \put(2, 56.4 ){\color{black} E}
        \put(2, 18 ){\color{black} F}
    \end{overpic}

    \caption{Autoencoder performance. A. No tracks in the image. B. One track in the image. C. Two tracks in the image. D. Three tracks in the image. Notice third track in left bottom corner, successfully detected by the \ATTI{} model. E. Four tracks in the image. Models have learned to represent confidence in track presence as distance from view center rather than the latent variable $a$. F. With $\geq$4 tracks in the image the models start to partially fail. \newline Leftmost column shows input images. For each model the image output of the decoder and the lines drawn according to the latent representation $z$ are shown. 
    In the input and output columns color depicts brightness. In track columns: green and blue lines correspond to enabled ($a>0$) and disabled ($a \leq 0$) tracks; for models without parameter $a$ all tracks are shown in green; crosses of corresponding color show the predicted $x,y$ position; white frame shows span of the input image. (Best seen in electronic version.)}
    \label{fig:fig4}
\end{figure}
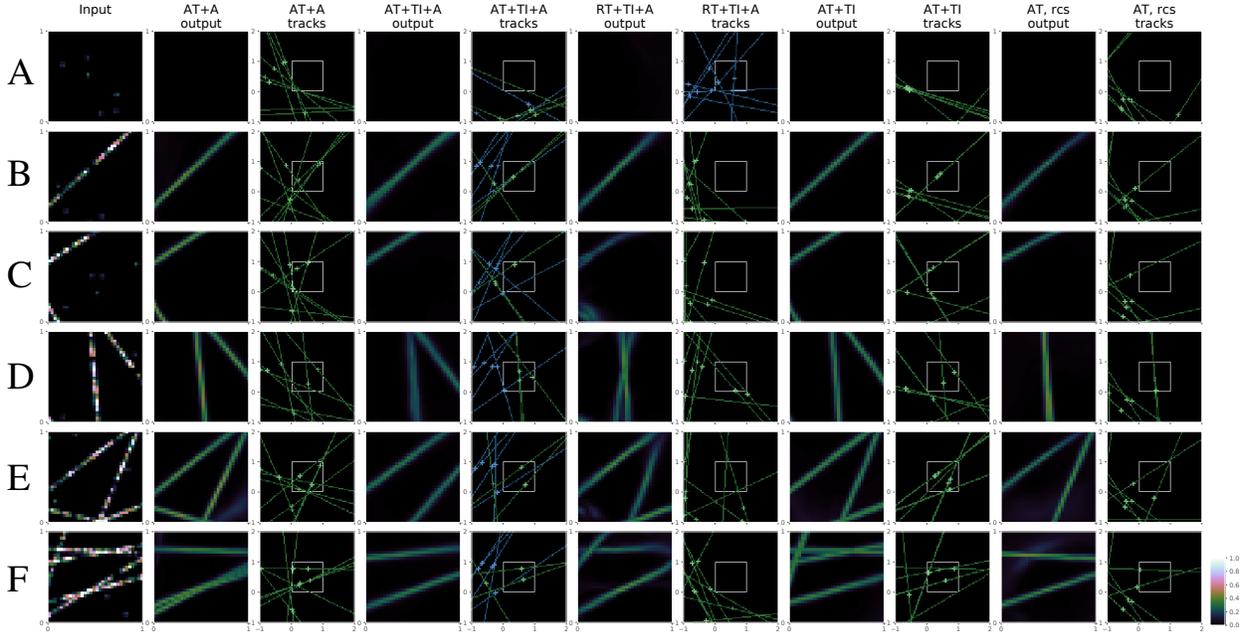

First, we evaluate whether our models have learned to properly capture the content of presented images in both latent and image spaces. In Figure \ref{fig:fig4}, the comparison of the outputs of the five models and the lines drawn according to the latent representation $z$ predicted for the image, interpreted as described above, are shown. It is clear that both \ATA{} and \ATTIA{} models were able to build the geometrically meaningful latent representation $z$ in most cases. 
For the \ATA{} model, which does not employ the translational invariance, the output contains more false detection both in the image output and in the drawn track lines. Also, the image output is significantly less sharp in the beginning of the training for this model. 
\RTTIA{} on the other hand did not manage to separate the factors of variation in the desired way, and it took much longer to converge, even just to mimic the desired image output. One can see inconsistency between the image output of the autoencoder and the track lines obtained by the latent representation, meaning that overall it did not grasp the concept of the geometrical space in the desired manner. None of the models properly learned the ability to “switch off” the tracks using the confidence parameter $a_i$. While this parameter is not completely ignored (blue lines in the tracks column in Fig. \ref{fig:fig4}), in most cases the models have learned other ways to disable track parameter containers $z_i$, which are not used to encode lines in the image. These containers simply have geometrical parameters corresponding to lines outside of the image crop range, or have $x,y$ coordinates far away from the image crop center (e.g.  \ATA{} and \ATTIA{} in Fig. \ref{fig:fig4}A). Performance of the \ATTI{} model was comparable or sometimes even better than that of the \ATTIA{} model. On the downside, without the parameter $a_i$ acting as a regularizer, this model tends to attribute close parameters to several lines (e.g. \ATTI{} in Fig. \ref{fig:fig4}E). The overparametrized models used the $x,y$ position to encode confidence in the track presence by placing them closer or further from the image along the track line (Figure \ref{fig:fig4}E,F). Performance of the \ATrcs{} model is slightly worse than of the \ATTI{} model. The performance of the models clearly degrades when the number of tracks in the image crop is $\geq$4. We assume that the main reason for this is that these cases were rather underrepresented in the training set.

\subsection{Disentanglement of the geometrical variational factors}
\begin{figure}
\centering
    \includegraphics[width=\textwidth]{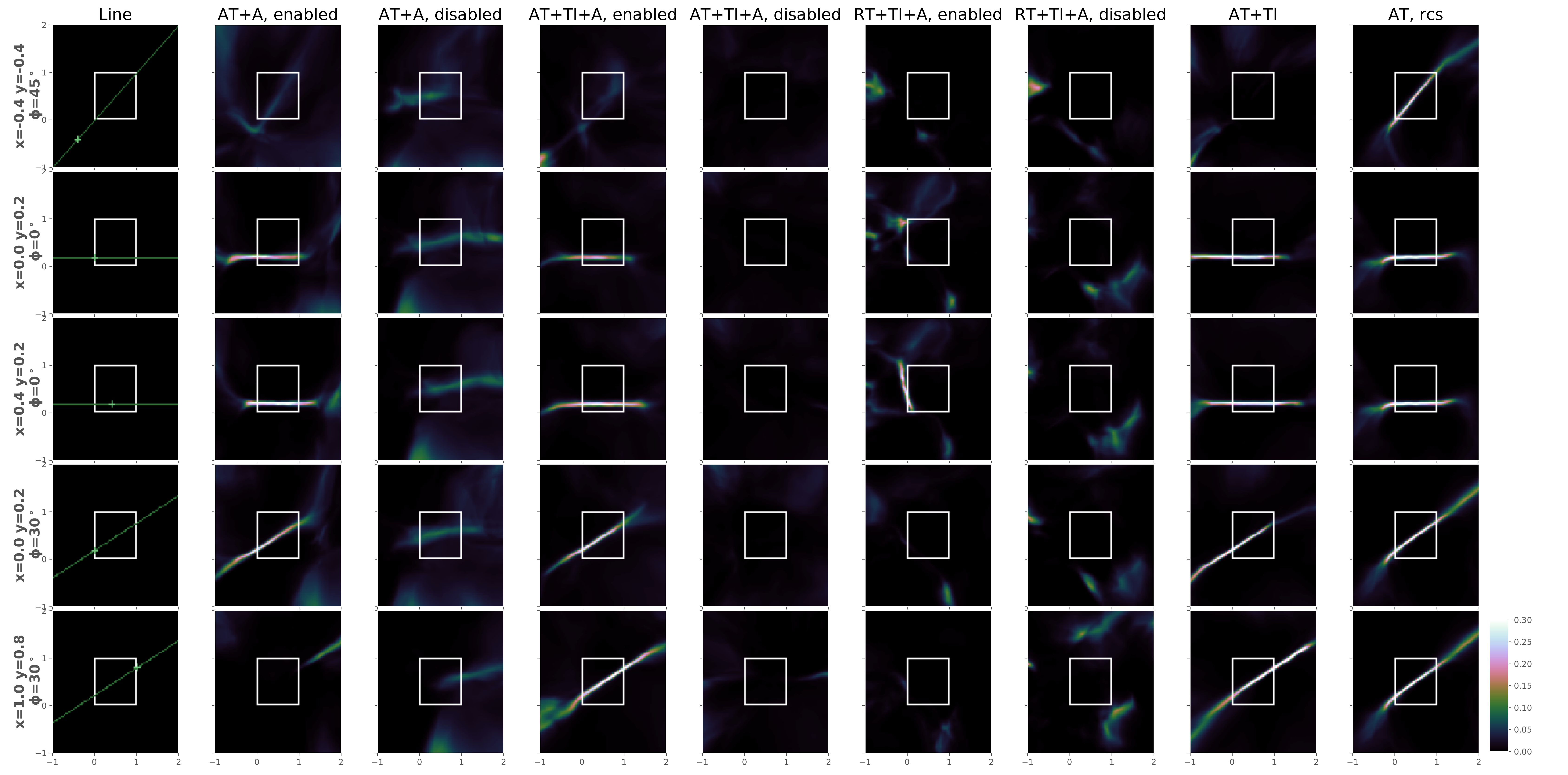}

    \caption{Comparison of decoder outputs for all models for given latent variable values. Tracks corresponding to the given values are shown in leftmost column. For disabled and enabled tracks, the activation parameter is set to $a=0$ and $a=1$ correspondingly. (Best seen in electronic version.)}
    \label{fig:fig5}
\end{figure}

To better understand the learned representation, we have performed a careful dissection into both decoder and encoder in this and the following sections, that was possible since the latent representation was designed to be fully interpretable. We start with the visual analysis of the learned representation by verifying the output of the decoder for given values of $z_i$. To this end, we have run the decoder on the entire range of meaningful values of $z_i$. In addition, for this study we performed the prediction on the image coordinate area $(-1,-1)-(2,2)$, i.e. 9 times bigger than the range of the original cropped image $(0,0)-(1,1)$. This way we can empirically see how well the decoder generalizes to a wider coordinate range. This is possible thanks to the CoordConv nature of the decoder: by changing the values in the coordinate channels we can perform the prediction at any position. In Figure \ref{fig:fig5} and Supplementary Figure \ref{sfig:sfig1} we show comparisons of decoder’s predictions for a set of representative values of $z_i$ between all models.

It is clear that all models, which employ the activation parameter, have learned to suppress the output when the values of the parameters $a_i$ are small. The output of the \RTTIA{} model does not correlate with the expectation at all, and while the output resembles lines, the learned representation is clearly not the desired one. It would be interesting to investigate which representation was found but this is outside the scope of this paper. Models employing translational invariance produced more elongated lines that fade out slower compared to the \ATA{} model (compare, for example, \ATA{} vs \ATTIA{} and \ATTI{} in Fig. \ref{fig:fig5}, rows 2-5). \ATTI{} shows even more pronounced and fine lines. All overparametrized models (i.e. all except \ATrcs{}) also suppress tracks with an $x,y$ position lying further from the image range (Fig. \ref{fig:fig5}, top row).

\subsection{Performance of the track parameters’ measurement}
\label{sec:traperf}
\begin{figure}
    \includegraphics[width=0.91\textwidth]{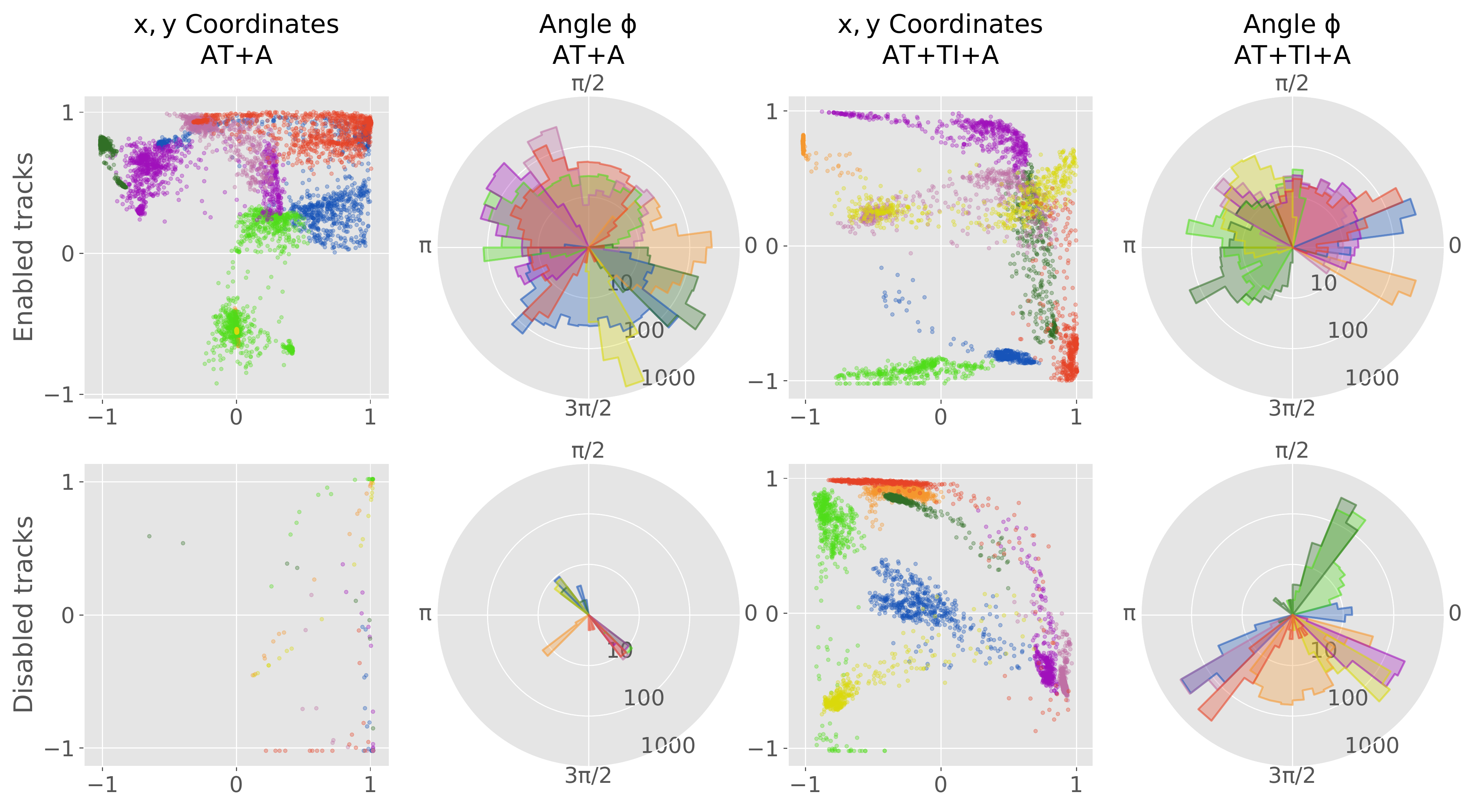}
    \includegraphics[trim=0 0 150 0,clip,width=0.99\textwidth]{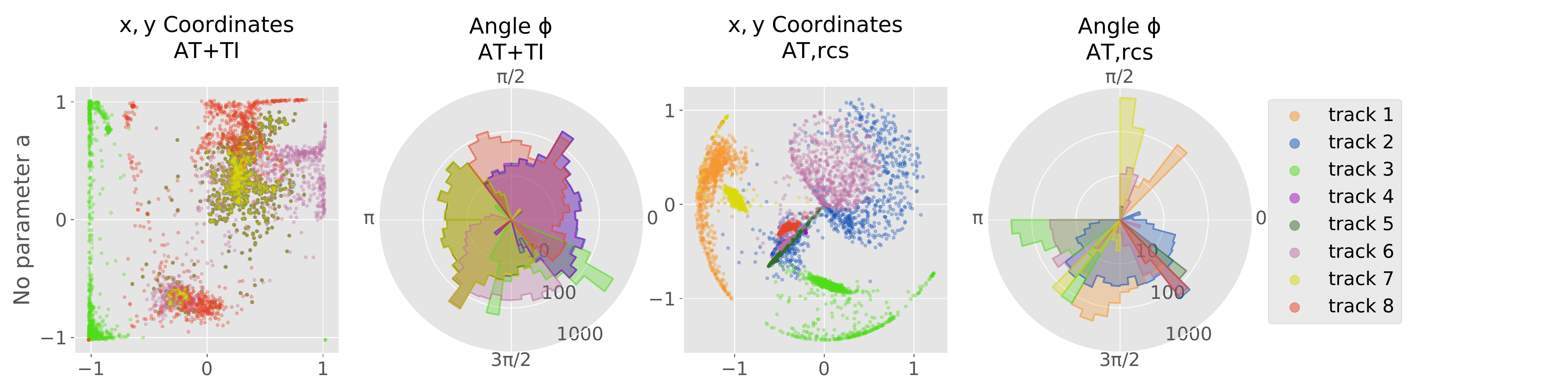}

    \caption{Distribution of the $x,y$ coordinates and inclination angle $\phi$ for the enabled (top row) and disabled (middle row) tracks for the \ATA{} and \ATTIA{} models and the models without activation parameter \ATTI{}, \ATrcs{} (bottom row). Interpreted geometrical values for the eight track parameter containers are shown in different colors. Values for each container tend to cluster, covering a particular space region.}
    \label{fig:fig6}
\end{figure}

To study the performance of the model for tracking, we evaluate the distribution and resolution of the encoder outputs $z_i$. In Figure \ref{fig:fig6}, the distributions of the predicted $x_i,y_i$ positions and the angle $\phi_i$ obtained from the $c_i,s_i$ parameters for each of the eight track feature containers, $z_i$, is shown for \ATA{}, \ATTIA{}, \ATTI{}, and \ATrcs{} models. Since the latter has a different representation, for consistency we obtain the values of $x_i,y_i,\phi_i$ using the two-argument arctangent function as follows:
$$\theta_i=\arctan2(s_i, c_i)$$
$$x_i=r_i\cos \theta_i; y_i=r_i \sin\theta_i; \phi_i=\theta_i-\pi/2$$

We skip further studies of the \RTTIA{} model, since the latent variables do not have the desired meaning, driving the geometrical analysis meaningless. We show these distributions separately for “enabled” and “disabled” tracks, according to the latent activation parameter $a_i$, where applicable. One can clearly see again that the models did not learn to use the parameter $a_i$, and e.g. the \ATA{} assigned almost all tracks the “enabled” value of the activation parameter. 
Instead, the reconstructed parameters for containers, which do not correspond to any tracks in the image, have rather localized $x,y$ positions and angle (see the peaks in the angular distribution, present for each of the eight track containers).
The positions for existing tracks lie within or close to the coordinate region of the image $(0,0)-(1,1)$. While they tend to cluster for each track container, they rather uniformly cover a wide band in the $x,y$ space. Notably, combined with a wide angular distribution, this localization does not limit the sensitivity region of the model (see, for example, the parameter distributions of the 8-th container for the enabled tracks in the \ATA{} model in Fig. \ref{fig:fig6}).

In the angular space all directions are covered, leaving no blind spots. Each of the eight parameter containers covers a subspace with some overlap for models \ATA{} and \ATTIA{}. This means, that only a fraction of the track containers is sensitive to any chosen direction. E.g. the \ATTIA{} model would fail to detect >3 parallel lines with $70^{\circ}$ inclination. Angular overlap in the \ATA{} model is very poor leading to poor detection of several parallel tracks in an image crop. In fact, three out of eight output containers have geometrical parameters corresponding to lines outside of the image (Supplementary Figure \ref{sfig:sfig2}). In \ATTI{}, on the other hand, each container covers almost $\pi$ in angular space, and the overall distribution is rather uniform (See Supplementary Figure \ref{sfig:sfig2}). This would allow to detect several parallel lines in a view (e.g. in Fig. \ref{fig:fig4}F several tracks have similar angles).

The \ATrcs{} model did not learn to utilize most of the containers. Practically only the containers 2 and 6 learned to encode track lines, as seen in Figure \ref{fig:fig6}. Nevertheless, even these two containers do not cover the whole angular range. For the remaining containers, the tracks lie outside of the image crop region, as seen on the $x,y$ distribution. This distribution is easy to interpret for this model since the track angle can be observed directly from the coordinates. For a circle with center at the origin and passing through some point $x,y$, the track line would be the tangent to the circle at this point. Arguably, the lack of flexibility due to minimal parametrization did not allow this model to efficiently switch off the tracks, leaving 2 almost always enabled and 6 always disabled.

To quantitatively evaluate the performance of the models we have processed the test dataset. This dataset was generated similarly to the training dataset with additional information on ground truth (GT) track positions and angles. It consists of 30,000 images with 5,000 sample images for each of 0, 1, ..., 5 tracks/image conditions. 

First, we assign the reconstructed tracks (and active, i.e. $a>0$ for models with activation parameter $a$) to the GT ones or mark them as fake. We evaluate the distance $\Delta r$ from image center between a predicted track and a GT track, and the difference in angle $\Delta \phi$. Then we build the $\chi^2$ as $\chi^2=(\frac{\Delta r}{\sigma_r} )^2+(\frac{\Delta \phi}{\sigma_\phi})^2$. Here we use the theoretical position resolution which is defined by pixel size $\sigma_r=\frac{1 px}{\sqrt{12}}=\frac{1}{32} \frac{1}{2 \sqrt{3}} \approx 0.009 \approx 0.3$ px, and angular resolution defined by pixel size and track length $l$ within the image crop $\sigma_\phi=\frac{\sqrt{2}}{\sqrt{12}l}=\frac{1}{\sqrt{6}l}$, ($\sigma_\phi \approx 13$ mrad for $l=32$ px).

\begin{figure}
    \includegraphics[width=\textwidth]{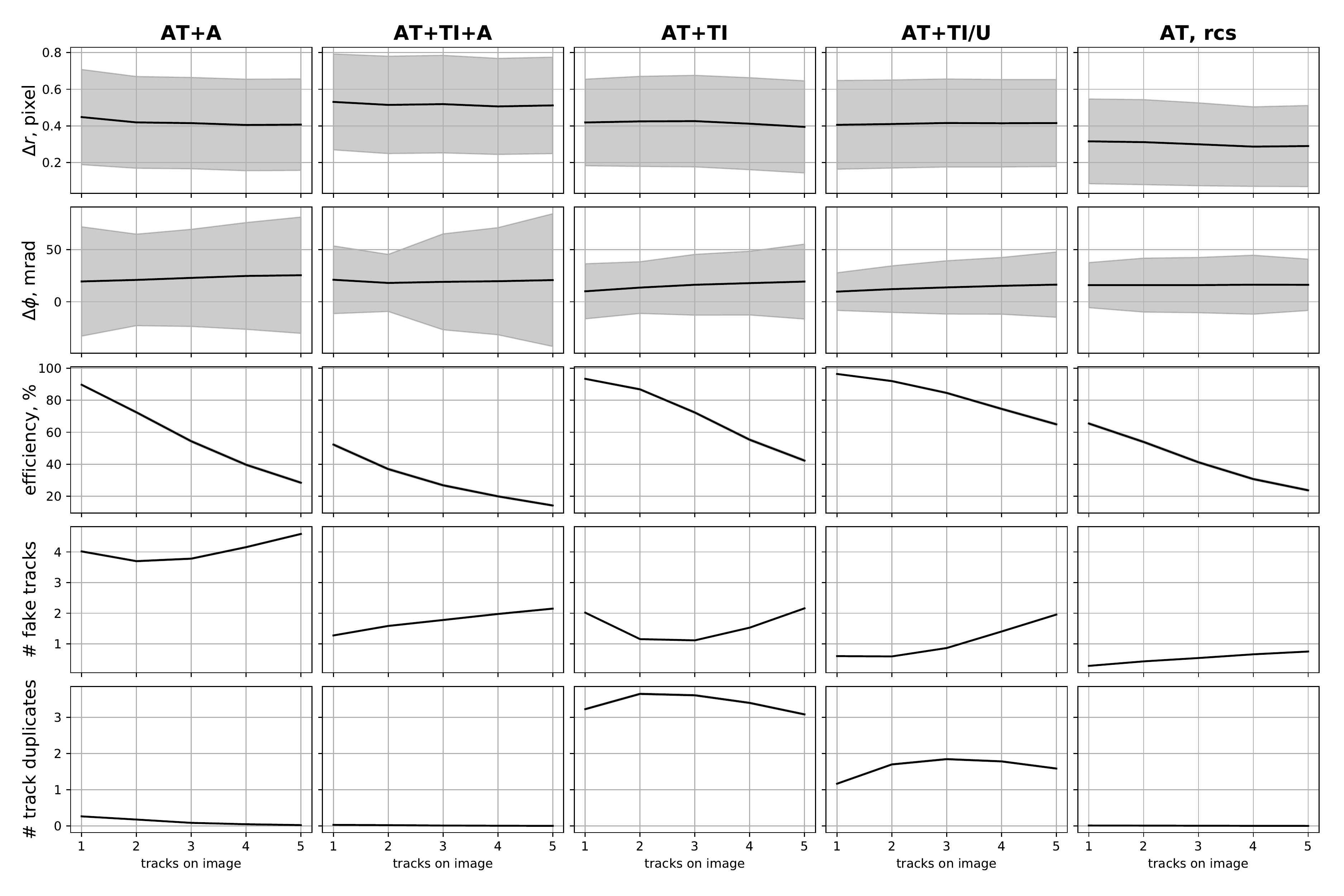}
    
    \caption{Tracking precision and efficiency as function of the number of tracks in the image for the five model configurations. From top to bottom: distance between GT track and the reconstructed one (less is better); angular difference (less is better); fraction of reconstructed tracks (more is better); number of reconstructed tracks that have no corresponding GT track, per image (less is better); number of reconstructed tracks that duplicate another reconstructed track, per image (less is better). Gray fill shows error range of 1 standard deviation of distribution for resolution parameters and of mean value for efficiency, fake tracks, and duplicates.}
    \label{fig:fig7}
\end{figure}

The assignment is then performed sequentially, by selecting available prediction-GT pairs according to the minimum value of $\chi^2$, if $\chi^2 \leq 11.83$  ($3\sigma$ statistical significance, number of degrees of freedom $ndf=2$). The remaining predicted tracks are split into two categories, fakes and duplicates. A track is considered as a duplicate if its $\chi^2$ to any of the used GT tracks or assigned prediction tracks is $\chi^2<2.3$ (i.e. within $1\sigma$), and as a fake otherwise. For the assigned tracks, we then evaluate the offsets $\Delta r$, $\Delta \phi$ as a function of number of tracks on the original image, as well as the fraction of assigned tracks (efficiency), number of fake tracks, and number of duplicate tracks (Figure \ref{fig:fig7}). The actual coordinate and angular resolutions $\sigma_{r,mod}$, $\sigma_{\phi,mod}$ of the models can be estimated from these data as mean values of $\Delta r$ and $\Delta \phi$. For example, for the \ATTI{} model $\sigma_{r,mod} \approx 0.013 \approx 0.42$ px, $\sigma_{\phi,mod}\approx 15$ mrad for $l=32$ px).

We then use mean resolutions for each model to show the 
$\chi^2=(\frac{\Delta r}{\sigma_{r,mod}} )^2+(\frac{\Delta \phi}{\sigma_{\phi, mod}})^2$ distribution for these models for different numbers of tracks per image crop in Supplementary Figure \ref{sfig:sfig3}. While for the \ATA{}, \ATTI{}, and \ATrcs{} models the distributions are consistent with a $\chi^2$ distribution with $ndf=2$, for the \ATTIA{} model, the peak is smeared and shifted towards higher values, consistent with higher resolution variance especially in the high track density region.

The resolution of the models is stable as the number of tracks grows. The \ATTI{} model has consistently higher resolution, as well as higher efficiency. Efficiency significantly decreases in all models with increasing number of tracks per image. We argue that it is caused by the fact that images with high track number were underrepresented in the training set (Figure \ref{fig:fig2}C), and the efficiency would improve if a training set with high track multiplicity images were used. To support this claim, we have generated a training dataset of 60,000 images with a uniform distribution of track density, i.e. 10,000 images for 0, 1, ..., 5 tracks/image crop. We have then retrained the \ATTI{} model on this dataset. This has significantly (>20\%) improved the efficiency for high track density (\ATTIU{} model in Fig. \ref{fig:fig7}).
While the number of fake tracks is similar in the \ATTI{} model, since it lacks the regularization based on the latent parameter $a$, it tends to assign all of the available containers to tracks. This leads in turn to a larger number of duplicates. Nevertheless, this effect is suppressed for the \ATTIU{} model, trained on the dataset with uniform track number representation. Another model without the activation parameter $a$ – the \ATrcs{} model does not produce many duplicates, most likely since each track is sensitive only to a narrow angular range, as shown above.

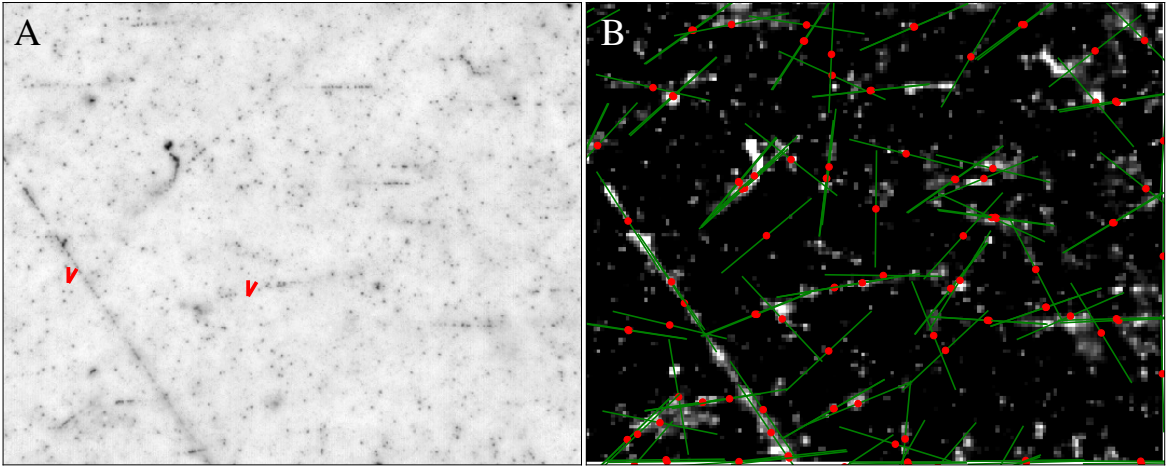
\begin{figure}
    \begin{overpic}[width=\textwidth]{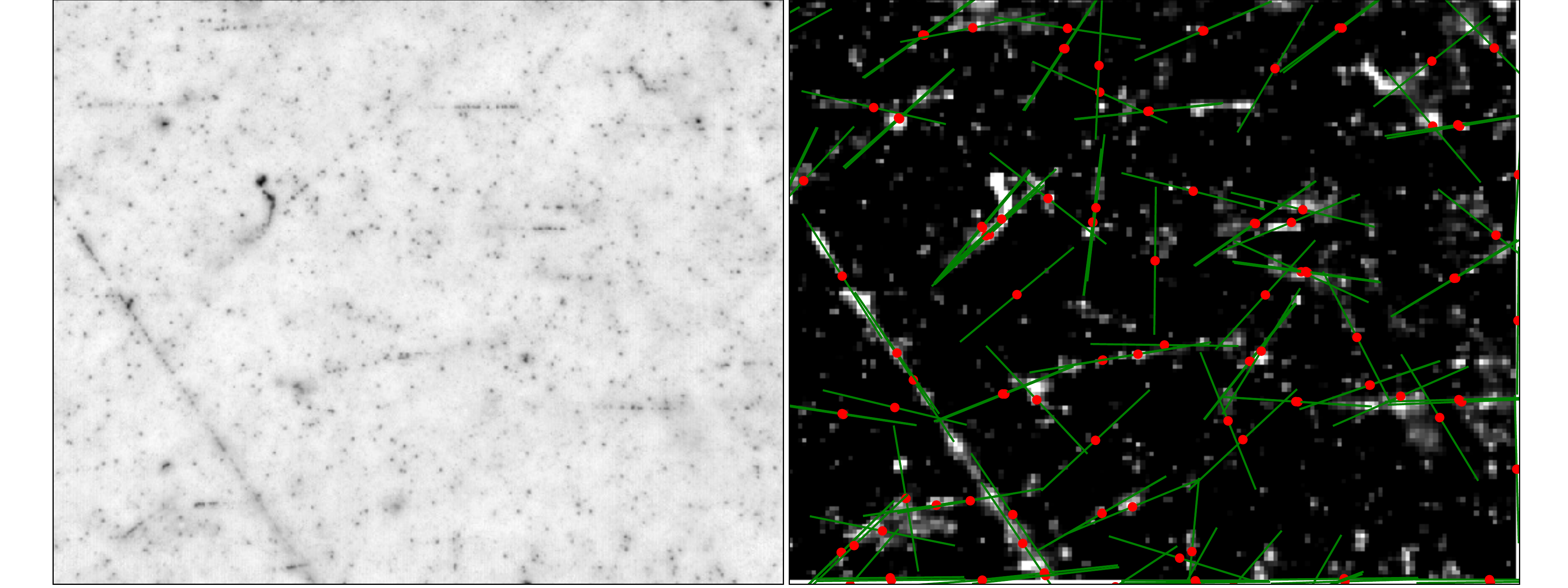}  
        \Large
        \put(20,    160){\color{black} A}
        \put(242,   160){\color{white} B}
        
        \linethickness{0.5mm} \color{red}
        \put(37,70){\line(0,5){6}\line(0.3,0.5){3.2}}
        \put(105,65){\line(0,5){6}\line(0.3,0.5){3.2}}
    \end{overpic}
    
    \caption{Tracking particles in emulsion data with the \ATTIU{} model. A. Wide field microscopy image of an emulsion detector. The image corresponds to a $380 \mu m\times300 \mu m$ detector surface. B. A downscaled and inverted image, overlaid with track segments (32 pixels long) reconstructed with the \ATTIU{} model. Examples of tracks reconstructed consistently with our model are highlighted by arrowheads.}
    \label{fig:fig8}
\end{figure}

Finally, we processed a real emulsion dataset to qualitatively observe the performance of the \ATTIU{} model (trained on synthetic data) in processing real experimental data. We used a single image out of a 3D tomographic image stack of size $640\times512$ pixels corresponding to $190\times150\,\mu m$ of emulsion detector area, irradiated with 400 GeV protons at different angles at the SPS accelerator beam at CERN \cite{Aoki2019}. We preprocessed the image (Figure \ref{fig:fig8}A) by downscaling it by a factor of four, inverting the image, and normalizing the color scale to match the training data properties (Figure \ref{fig:fig8}B). We then divided the image into 5$\times$4 non-overlapping 32$\times$32 pixel crops and processed them independently. The resulting tracks are then assembled into the full image size and shown as 32 pixels long segments with highlighted $x,y$ position (Figure \ref{fig:fig8}B, overlay). Even though the models were trained on synthetic data with different signal and noise distributions from experimental data, one can appreciate the agreement between real tracks and predicted ones (e.g. tracks pointed to by arrowheads), confirming the effectiveness of our method. For real-life applications, the model must be trained on the raw experimental data, to learn the true signal and noise distributions from it.

\section{Discussion and outlook}
\label{sec:Discussion}
Disentangling factors of variation remains a hot topic for several years in representation learning research. Moreover, developing models that are capable of abstracting high-level concepts from raw data can lead to plenty of direct practical applications. Many works approached this problem by employing variational autoencoders with regularizations in the latent representation that enforce disentangling \cite{Kumar2017} or autoencoders combined with adversarial training \cite{Szabo2018}. In most cases, after disentangling the factors of variation, a few labeled samples can be used to associate the factors with interpretable measures in a quantitative way. Previously it has been shown \cite{Locatello2019} how a few labeled samples can improve disentanglement itself. In another work \cite{Kim2018} it was shown that applying an equivariance constraint, i.e. changing one factor of variation, corresponding to a change of one dimension of the disentangled representation in a predictable manner, leads to disentangled variables. Nevertheless, to the best of our knowledge no previous works have tried to extract meaningful quantitative information in a fully unsupervised manner.

In this work, we have demonstrated that imposing equivariance constraints on the autoencoder under geometrical transformations in the image and latent representation domains enables the model to “discover” the existence of multiple lines in the presented images in a fully unsupervised manner. Incorporating simple affine transformation such as translation, rotation, scaling and skew as equivariances between the image and latent spaces allows the models to successfully disentangle the factors of variation in the image data into geometrically meaningful parameters (coordinates and angles of lines). Adding the possibility to “switch off” a predicted track with an additional activation parameter $a$ does not drastically change the results (models \ATA{} and \ATTIA{}). While it can help to prevent the shortcut problem and reduces the number of track duplicates, these models did not learn to exploit it.

Incorporating the translation along the track line in addition to the whole set of affine transformations enforces the line detection. However, employing only the translational invariance, even together with rotational transformations (\RTTIA{} model), leads to reference ambiguity so that latent parameters do not correspond to the desired geometrical variables. We believe that with a few calibration measurements it would be possible to find a mapping from these latent parameters to the desired geometrical variables, but this lies out of the scope of this work. As we have shown, a larger set of transformations allows the model to immediately learn the latent representation in an unambiguous, geometrically meaningful way. The minimal subset of affine transformations, sufficient for disentangling the factors of variation without reference ambiguity, will be explored in future works.

In addition to the coordinate-angle parametrization, which gives the models more freedom in sample exploration, we have studied the classical rho-theta parametrization. Under the set of affine transformations, this model is also able to learn a meaningful parametrization in a fully unsupervised manner. Yet the model performance is slightly worse than, for example, the \ATTI{} model, most likely because the incorporated CoordConv approach prefers to have a natural $x-y$ coordinate representation.

The main weak point of our current implementation is that neither the background nor the grain distribution along the lines are in any way represented by the current models, which may impair line detection in the case of high background rate. In addition, the employed transformations in the image domain affect the image parameter distributions, such as brightness (corresponding to the $dE/dx$ energy loss in emulsion detector) and sharpness. Models with additional global and per-track latent parameters without an \textit{a priori} assigned meaning would naturally overcome these hurdles. Training them would require dealing with the shortcut problem in these parameters, and thus would benefit from employing the adversarial framework \cite{Szabo2018, Goodfellow2014}. While the aim of this work was to carefully study the proposed approach in general, we leave this aspect to further studies.

We expect this approach to have a large potential in the analysis and extraction of geometrical properties from image data. In further work we plan to adapt this technique to the location of tracks in full resolution 3D tomographic microscopy data, or data from Liquid Argon Time Projection Chamber detectors \cite{MicroBooNEcollaboration2018, Brailsford2018}, which would be a direct extension of this approach. Also adding more samples with a higher track number in the training dataset is expected to improve the efficiency and resolution in cases with high track density.

While designed to detect simple line structures, this technique has the potential to be used for locating and parametrizing other objects, such as splines. This would enable the tracking of particles in magnetic fields and pave the way to novel automated image vectorization techniques. Being fully unsupervised, this approach can leverage all available raw dataset with no extra work required.

\section*{Acknowledgement}
The authors would like to thank Paolo Favaro and Attila Szab\'{o} for fruitful discussions.
\printbibliography
\newpage
\appendix
\section{Affine transformations in latent representations}
\label{app:Transformations}
In this section, we describe the implementation of the affine transformations for the employed $(x,y,c,s)$ and $(r,c,s)$ representations. Additionally, we provide the parametrization of the transformations themselves, as well as the range of parameters used during training in this study.
We produce the transformation functions in the latent and image spaces as a combination of rotation, scaling, skew, and shift. First, we define the corresponding transformation operations of the track line parameters in the following way (index $i$ is omitted for brevity).
Rotation:
$$T_{rot} (z_t \vert \xi_r )=(x,y,\cos(\phi+\xi_r ),\sin(\phi+\xi_r)); \phi=\arctan2(s,c); \xi_r\in (-\frac{\pi}{4}, \frac{\pi}{4})$$
where $\arctan2(y,x)$ is the two-argument arctangent function.
Scaling:
$$T_{scale_x} (z_t \vert \xi_{sc_x} ) = (\xi_{sc_x} x,y,\frac{\xi_{sc_x} c}{\epsilon+\sqrt{(\xi_{sc_x} c)^2+s^2 }}, \frac{s}{\epsilon+\sqrt{(\xi_{sc_x} c)^2+s^2}}); \xi_{sc_x} \in (0.7,1.3); \epsilon=10^{-6}$$
$$T_{scale_y} (z_t \vert \xi_{sc_y } )= (x,\xi_(sc_y ) y,\frac{c}{\epsilon+\sqrt{c^2+(\xi_{sc_y}s)}^2 },\frac{\xi_{sc_y} s}{\epsilon+\sqrt{c^2+(\xi_{sc_y}s)^2}}); \xi_{sc_y} \in (0.7,1.3); \epsilon=10^{-6}$$
Skew:
$$T_{skew_x} (z_t \vert \xi_{sk_x} )=(x+\xi_{sk_x} y,y,\cos \phi,\sin \phi); \phi=\arctan2(s,c+\xi_{sk_x} s); \xi_{sk_x} \in (-0.4,0.4)$$
$$T_{skew_y } (z_t \vert \xi_{sk_y} )=(x,y+\xi_{sk_y} x,\cos \phi,sin \phi); \phi=\arctan2(s+\xi_{sk_y} s,c); \xi_{sk_y} \in (-0.4,0.4)$$
Translation:
$$T_{trans} (z_t \vert \xi_{t_x},\xi_{t_y} )= (x+\xi_{t_x},y+\xi_{t_y},c,s), \xi_{t_x} ),\xi_{t_x} \in (-0.4,0.4)$$
Rotation is performed around the coordinate origin; scaling $x$ and scaling $y$ preserves the $y$ and $x$ coordinates intact; skew $x$ and skew $y$ preserves points on the $x$ and $y$ axes  correspondingly. The employed range of transformations is a trade-off between urging the model to learn the desired representation and preserving most of the original tracks in the image crop after the transformation.

The combined space transformation is then produced by consecutively applying these transformations:
$$z'_t=T_{tr}(z_t \vert \xi)=T_{trans} (T_{skew_y} (T_{skew_x} (T_{scale_y} (T_{scale_x} (T_{rot} (z_t \vert \xi_r ) \vert \xi_{sc_x} ) \vert \xi_{sc_y} ) \vert \xi_{sk_x} ) \vert \xi_{sk_y} ) \vert \xi_{t_x},\xi_{t_y} ),$$
$$\xi=(\xi_r,\xi_{sc_x},\xi_{sc_y},\xi_{sk_x},\xi_{sk_y},\xi_{t_x},\xi_{t_y} )$$
In some models, we add an additional transformation of the track parameters $z_t$, corresponding to translational invariance (TI) along the line. This is implemented as a random shift of the $x,y$ parameters along the line:
$$T_{t.i.} (z_t \vert r) = (x+r c,y+r s,c,s);\; r=rand(-0.5,0.5).$$

We apply these transformations only to the enabled tracks according to the value of $a$. For the disabled tracks, the parameters are set to random values in the $(-1,1)$ range, enforcing the decoder to learn to ignore disabled tracks:
$$z'= 
\begin{cases}
    (z'_t,\sigma(\gamma a)), & a >0 \\
    (r_1,r_2,r_3,r_4,\sigma(\gamma a)); \; r_i=rand(-1,1), & a \leq 0
\end{cases}$$
where $\gamma=20$ and the sigmoid function $\sigma(a)= \frac{1}{1+e^{-a}}$ is applied to the activation parameter $a$ for implementation reasons.

For the model employing the $(r,c,s)$ representation, we first transform the parameters to the $(x,y,c,s)$ representation as: $\theta = \arctan2(s,c)$, $\phi=\theta-\pi/2$, $x=r \cos \theta$; $y=r \sin \theta$; $c=\cos \phi$; $s=\sin \phi$, and then apply the shown above transformations. Afterwards, the inverse transformation to the $(r,c,s)$ representation is applied.

The parameter set $\xi$ is drawn from a uniform random distribution for each training sample on each iteration and is fed into the network along with the images. The input images $I$ of size 96$\times$96 pixels are cropped to 32$\times$32 as shown in Figure \ref{fig:fig1b} and fed as the network input $I_c$. The same input images $I$ are then elastically transformed with the transformation function $I' = T_{im} (I \vert \xi)$ using the same parameter set $\xi$. For $T_{im}$ we have employed the \texttt{tf.contrib.image.transform} function from the TensorFlow library \cite{tensorflow2015-whitepaper}. The origin of the transformations corresponds to pixel coordinates (48,48) in the input image, i.e. the lower bottom corner of the crop. After being cropped to 32$\times$32 pixels, the images are used as network output target $I'_c$. We use larger 96$\times$96 input images to ensure that the final crop after transformation does not contain regions outside of the input image. Examples of the space transformations are shown in the Supplementary Figure \ref{sfig:sfig4}.

\newpage
\section*{Supplementary Material}
\beginsupplement{}
\begin{figure} [h!]
\centering
    \includegraphics[width=\textwidth]{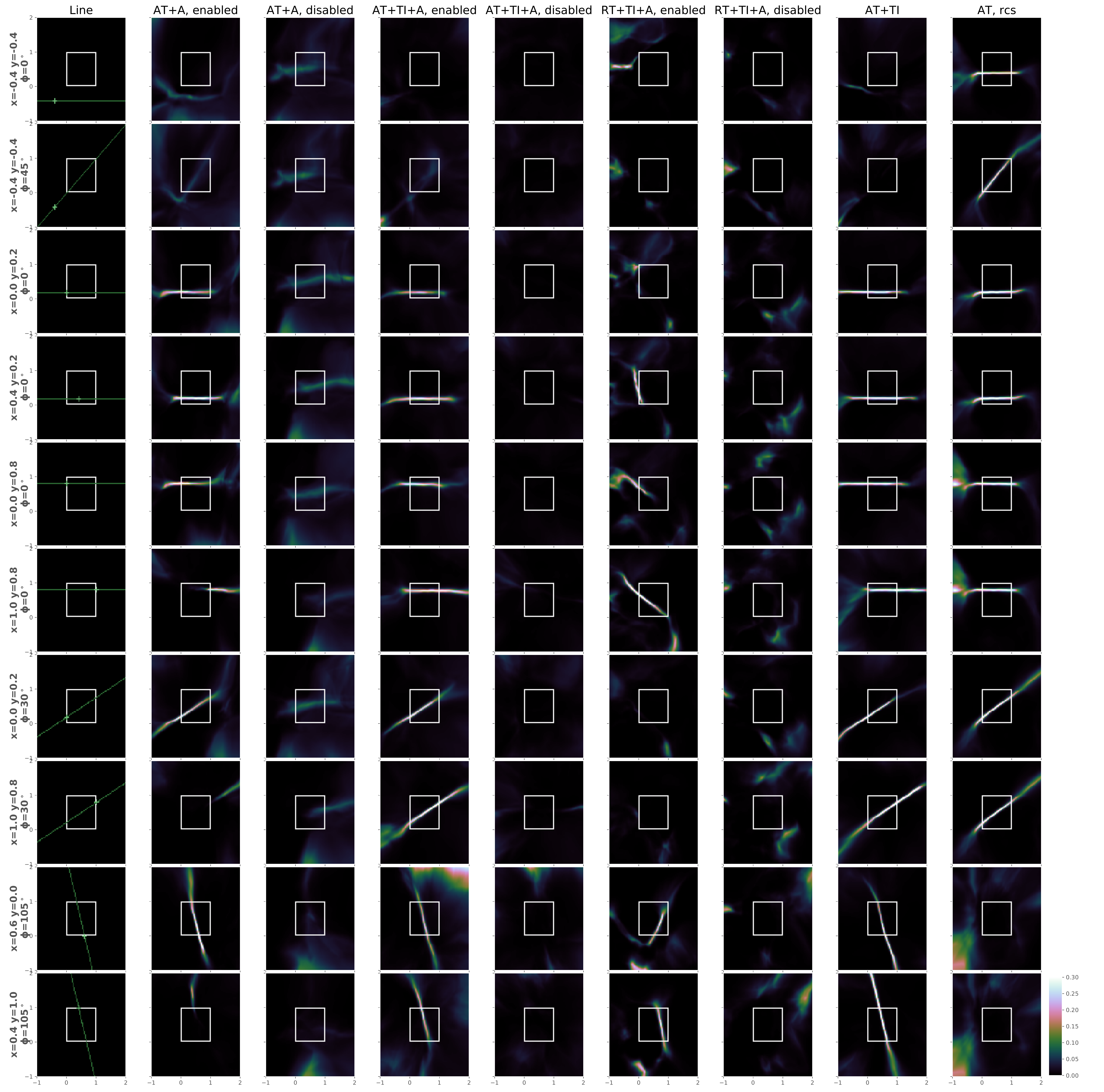}

    \caption{Comparison of decoder outputs for all models for several selected latent variable values. True tracks corresponding to the given values are shown in leftmost column.}
    \label{sfig:sfig1}
\end{figure}

\begin{figure}
\centering
    \begin{overpic}[width=0.6\textwidth]{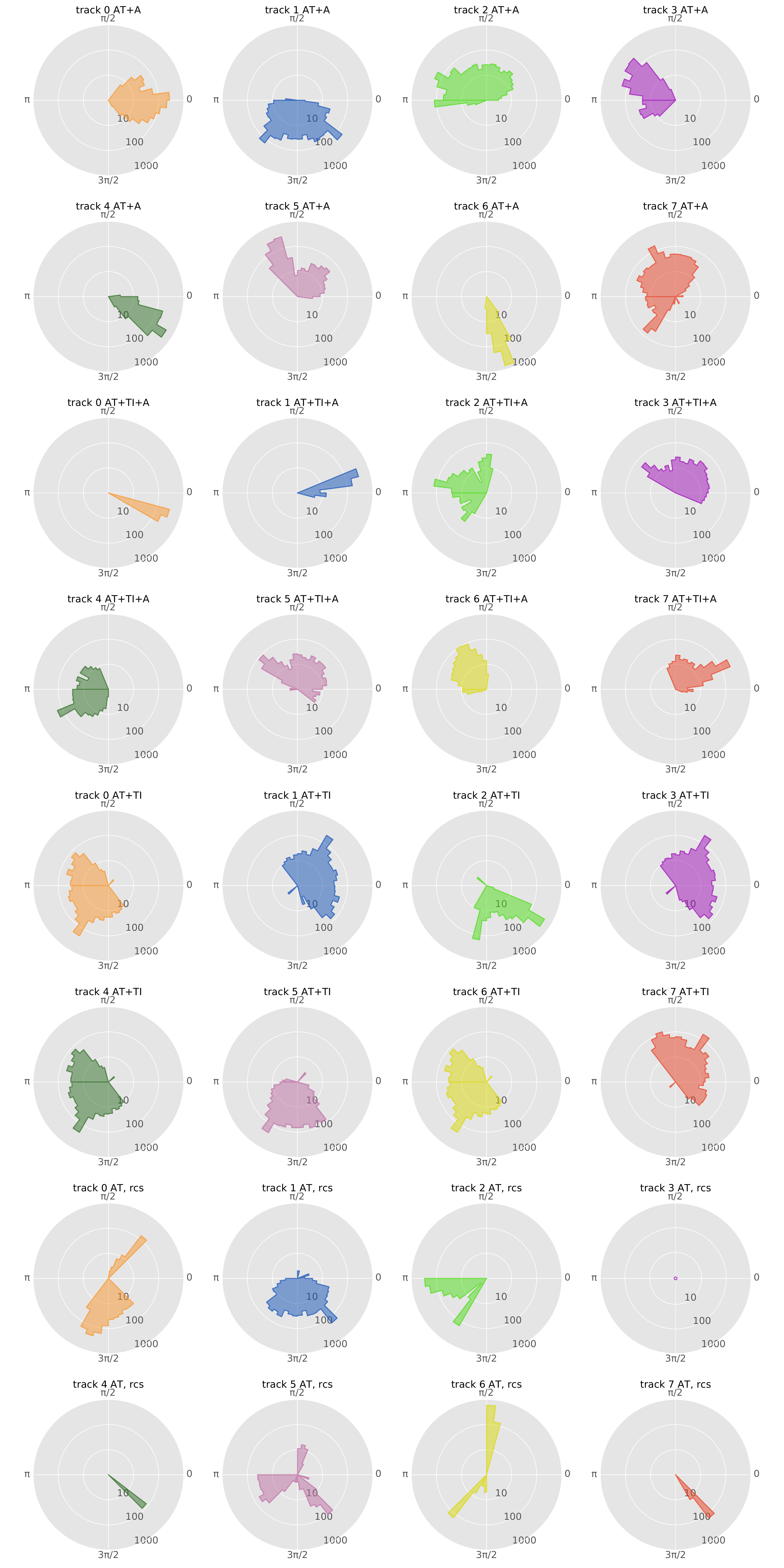} 
        \linethickness{0.25mm} \color{black}
        \put(0,423){\line(1,0){275}}
        \put(0,284){\line(1,0){275}}
        \put(0,141){\line(1,0){275}}
    \end{overpic}

    \caption{Distributions of the predicted track angles by each of the track feature containers in the latent representation. From top to bottom: \ATA{}, \ATTIA{}, \ATTI{}, and \ATrcs{} models. Color coding is the same as in Figure \ref{fig:fig6}.}
    \label{sfig:sfig2}
\end{figure}

\begin{figure}
\centering
    \includegraphics[width=\textwidth]{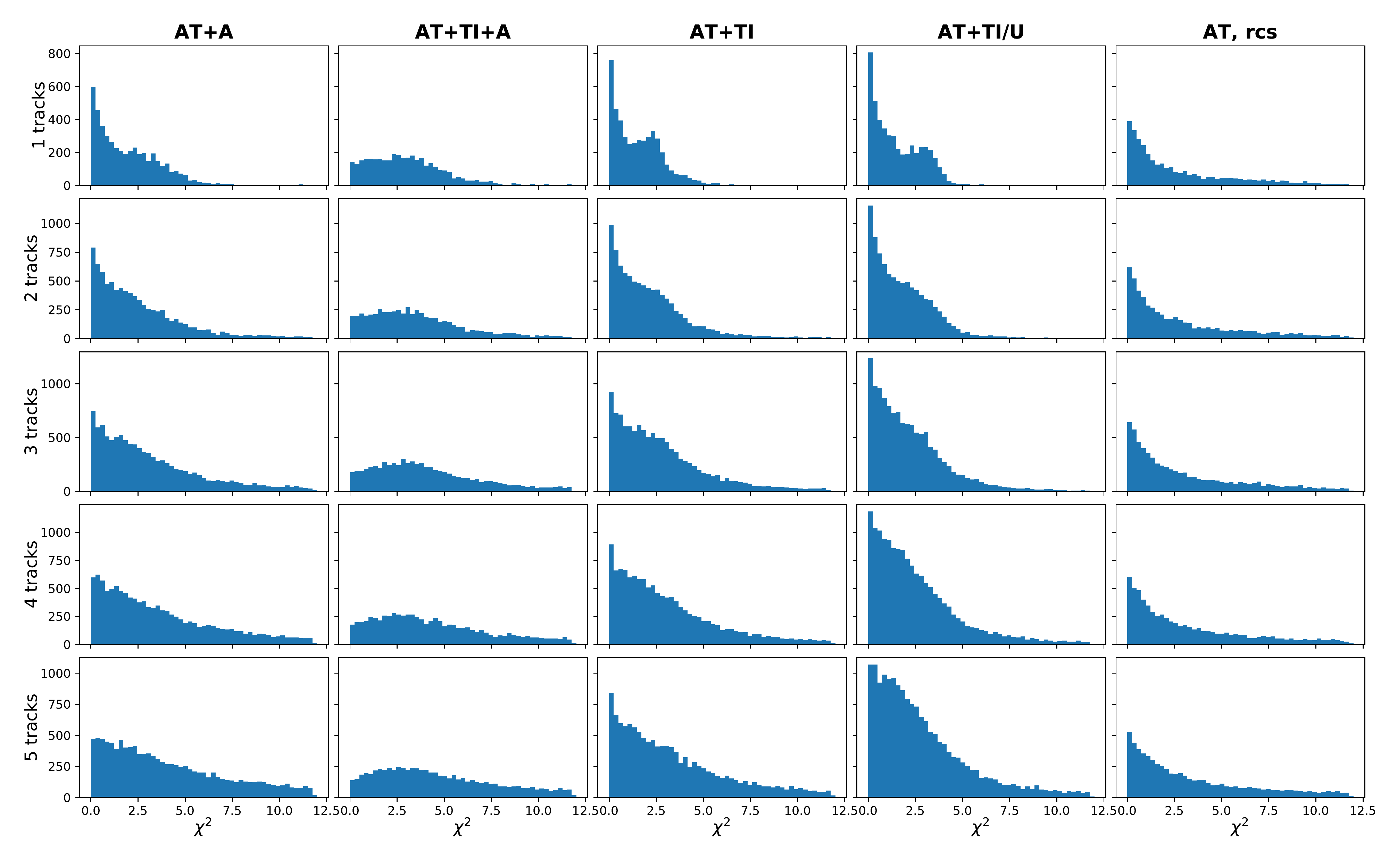}

    \caption{$\chi^2$ distributions for the \ATA{}, \ATTIA{}, \ATTI{}, \ATTIU{}, and \ATrcs{} models for 1, 2, 3, 4, and 5 tracks per image crop.}
    \label{sfig:sfig3}
\end{figure}

\begin{figure}
\centering
    \begin{overpic}[width=\textwidth]{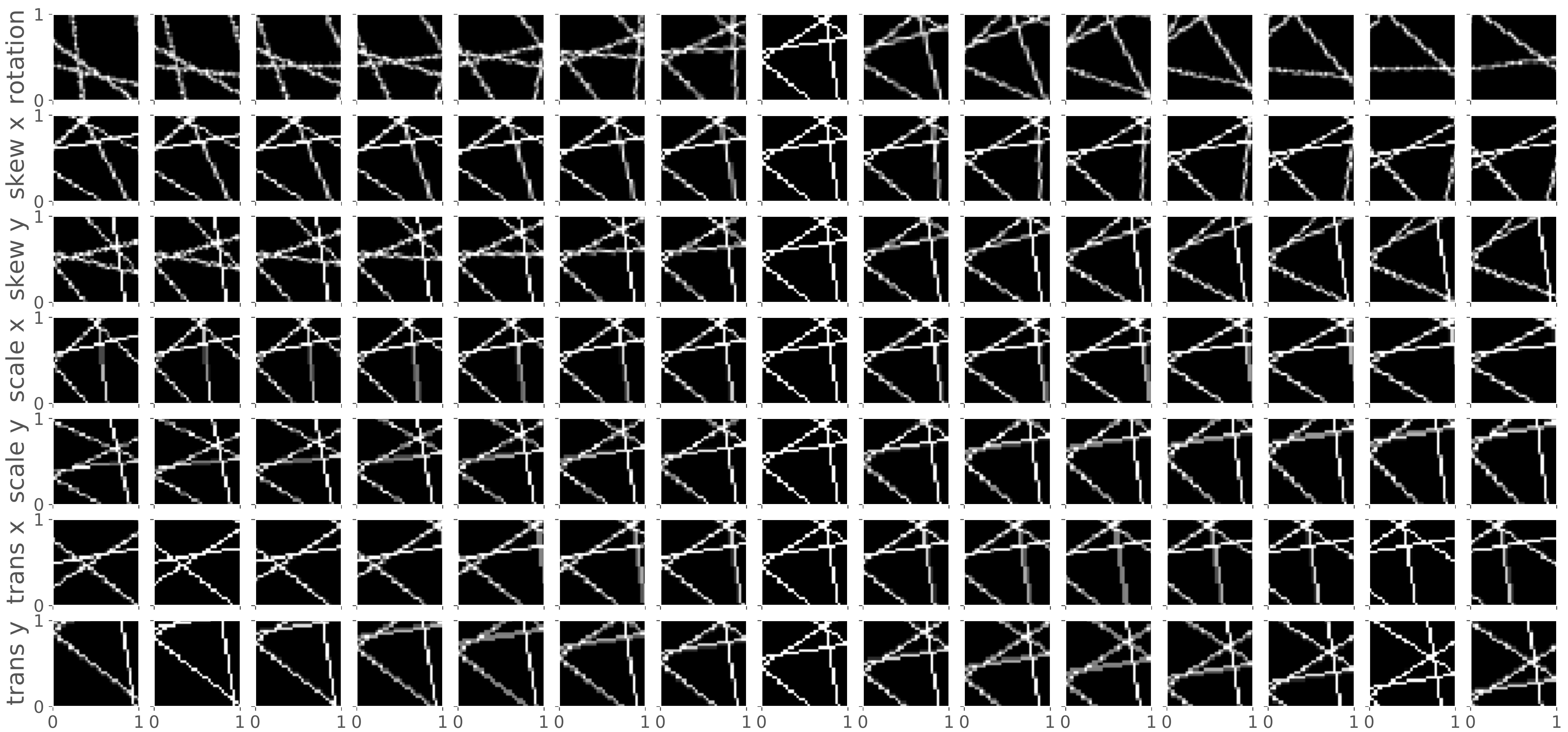} 
        \linethickness{0.5mm} \color{red}
        \put(0,0){\polygon(219,7)(248,7)(248,216)(219,216)}
    \end{overpic}

    \caption{Examples of image transformations (rows): rotation and skew, scale, and translation along x and y. \newline The transformations performed with 15 parameters (columns) are presented, linearly distributed in the employed parameter range. Transformations in the middle column correspond to identity transformations (highlighted).}
    \label{sfig:sfig4}
\end{figure}

\end{document}